\documentclass[11pt,a4paper]{article}
\usepackage{times}
\usepackage{latexsym}
\usepackage{amsmath}

\usepackage{graphicx}
\usepackage{emnlp2020}

\usepackage{soulutf8}
\usepackage{color}
\usepackage{microtype}
\aclfinalcopy 
\def\aclpaperid{423} 

\title{The RELX Dataset and Matching the Multilingual Blanks\\for Cross-Lingual Relation Classification}
\author{Abdullatif  K\"oksal \\
  Department of Computer Engineering \\
  Bo\u{g}azi\c{c}i University \\
  \texttt{abdullatif.koksal@boun.edu.tr} \\
  \And
  Arzucan \"Ozg\"ur \\
  Department of Computer Engineering \\
  Bo\u{g}azi\c{c}i University \\
  \texttt{arzucan.ozgur@boun.edu.tr} \\}

\date{}

\begin{document}
\maketitle
\begin{abstract}
Relation classification is one of the key topics in information extraction, which can be used to construct knowledge bases or to provide useful information for question answering. Current approaches for relation classification are mainly focused on the English language and require lots of training data with human annotations. Creating and annotating a large amount of training data for low-resource languages is impractical and expensive. To overcome this issue, we propose two cross-lingual relation classification models: a baseline model based on Multilingual BERT and a new multilingual pretraining setup, which significantly improves the baseline with distant supervision. For evaluation, we introduce a new public benchmark dataset for cross-lingual relation classification in English, French, German, Spanish, and Turkish, called \textbf{RELX}. We also provide the \textbf{RELX-Distant} dataset, which includes hundreds of thousands of sentences with relations from Wikipedia and Wikidata collected by distant supervision for these languages. Our code and data are available at: \url{https://github.com/boun-tabi/RELX}

\end{abstract}

\section{Introduction}
Extracting useful information from unstructured text is one of the most essential topics in Natural Language Processing (NLP). Relation classification can help achieving this objective by enabling the automatic construction of knowledge bases and by providing useful information for question answering models \cite{qa_rel}. Given an entity pair $(e1, e2)$ and a sentence $S$ that contains these entities, the goal of relation classification is to predict the relation $r \in R$ between $e1$ and $e2$ from a set of predefined relations, which may include 
`no relation' as well. For example, with the help of relation classification, we can create semantic triples such as 
\textit{(Rocky Mountain High School, founded, 1973)}
from a sentence like 
\textit{``Rocky Mountain High School opened at its current location in 1973 and was expanded in 1994.''}, where `Rocky Mountain High School' and `1973' are the given entities and `founded' is the relation between them based on this sample sentence.

Traditionally, relation classification methods rely on hand-crafted features \cite{handcraft1}. Lately, pretrained word embeddings \cite{word2vec} with RNN-LSTM architecture \cite{kbp37, lstm2} or transformers based models \cite{mtb} have gained more attention in this domain. Although non-English content on the web is estimated as over 40\% \cite{web_ratio} and the number of multilingual text-corpora is increasing \cite{text_mining}, recent studies on relation classification have generally focused on the English language. These supervised approaches for relation classification are not easily adaptable to other languages, since they require large annotated training datasets, which are both costly and time-consuming to create. 

The challenge of creating manually labeled training datasets for different languages can be alleviated through cross-lingual NLP approaches. In cross-lingual relation classification, the objective is to predict the relations in a sentence in a target language, while the model is trained with a dataset in a source language, which may be different from the target language. For example, a cross-lingual relation classification model should be able to extract semantic triples such as \textit{(CD Laredo, founded, 1927)} from a Spanish sentence like \textit{``CD Laredo fue fundado en 1927 con el nombre de Sociedad Deportiva Charlestón.\footnote{\textit{English Translation:} CD Laredo was founded in 1927 with the name ``Sociedad Deportiva Charlestón''.}''} for the given entities 
`CD Laredo' and `1927', even when the annotated training data is solely in English.

Thanks to multilingual pretrained transformer models like Multilingual BERT (mBERT) \cite{bert} and XLM \cite{xlm},  cross-lingual models have been studied in depth for several NLP tasks such as question answering \cite{xquad, xqa, xlm_roberta}, natural language inference \cite{xlm, xlm_roberta, bad_turkish}, and named entity recognition \cite{xlm_roberta}. 

In this paper, we first propose a baseline cross-lingual model for relation classification based on the pretrained mBERT model\footnote{\url{https://github.com/google-research/bert}}. Then, we introduce an approach called \textbf{Matching the Multilingual Blanks} to improve the relation classification ability of mBERT in different languages with the help of a considerable number of relation pairs collected by distant supervision. Prior works on cross-lingual relation classification use additional resources in the target language such as aligned corpora \cite{cross_second}, machine translation systems \cite{open_cross}, or bilingual dictionaries \cite{ibm}. Our mBERT baseline model does not require any additional resources in the target language. The Matching the Multilingual Blanks model improves mBERT by utilizing the already available Wikipedia and Wikidata resources with distant supervision.

We present two new datasets for cross-lingual relation classification, namely RELX and RELX-Distant. \textbf{RELX} has been developed by selecting a subset of the  commonly-used KBP-37 English relation classification dataset \cite{kbp37} and generating human translations and annotations 
in the French, German, Spanish, and Turkish languages. The resulting dataset contains $502$ parallel test sentences in five different languages with $37$ relation classes. To our knowledge, RELX is the first parallel relation classification dataset, which we believe will serve as a useful benchmark for evaluating cross-lingual relation classification methods. \textbf{RELX-Distant} is a multilingual relation classification dataset collected from Wikipedia and Wikidata through distant supervision for the aforementioned five languages. We gather from 50 thousand upto 800 thousand sentences, whose entities have been labeled by the editors of Wikipedia. The relations among these entities are extracted from Wikidata.

Our contributions can be summarized as follows:
\begin{enumerate}
\item We introduce the \textbf{RELX} dataset, a novel cross-lingual relation classification benchmark with 502 parallel sentences in English, French, German, Spanish, and Turkish.
\item To support distantly supervised models, we introduce the \textbf{RELX-Distant} dataset, which has hundreds of thousands of sentences with relations collected from Wikipedia and Wikidata for the mentioned five languages.
\item We first present a baseline mBERT model for cross-lingual relation classification and then, propose a novel multilingual distant supervision approach to improve the model.

\end{enumerate}

The rest of the paper is organized as follows. The related work is discussed in Section \ref{sec:related_word}. The details about the datasets are presented in Section \ref{sec:dataset}. Our mBERT baseline model and the Matching the Multilingual Blanks (MTMB) procedure are described in Section \ref{sec:methods}. The experimental results for the mBERT model and MTMB are presented in Section \ref{sec:results}. Finally, we draw conclusions and discuss future work in Section \ref{sec:conclusion}.

\section{Related Work}
\label{sec:related_word}
In monolingual relation classification, traditional methods generally depend on hand-crafted features \cite{handcraft1}. After the introduction of word embeddings \cite{word2vec, glove}, many relation classification models used pretrained word embeddings with the RNN \cite{kbp37,lstm2} or CNN \cite{cnnpop, cnn2} architectures. With the strong performance of transformer networks for various NLP tasks \cite{bert,xlm, elmo}, \citet{mtb} applied BERT with different representations to the relation classification task and showed the strength of it on several English datasets.

Before the introduction of multilingual transformers \cite{bert,xlm,xlm_roberta}, cross-lingual word embeddings have been widely used in zero-shot cross-lingual transfer with word embedding alignments for different tasks such as named entity recognition \cite{ner_wemb} and natural language inference \cite{nli_wemb}. This approach has also been utilized for cross-lingual relation classification  \cite{ibm}. However, recently,  multilingual deep transformers have attracted lots of attention in several cross-lingual tasks such as question answering \cite{xquad, xqa, xlm_roberta}, natural language inference \cite{xlm, xlm_roberta, bad_turkish}, and named entity recognition \cite{xlm_roberta}. To the best of our knowledge, we present the first transformer based approach for the task of cross-lingual relation classification. In addition, we introduce a multilingual distant supervision method to improve the baseline transformer model. \citet{mtb} use a similar approach for monolingual relation classification, called Matching the Blanks. For the pretraining process, they collect pairs of English sentences based on the shared entities, annotated by an entity linking system. On the other hand, we propose a multilingual approach that utilizes Wikipedia and Wikidata, which are already available for many languages and have been successfully used for tasks such as multilingual question answering \cite{abdou2019x} and named entity recognition \cite{mner}. 

Most cross-lingual relation classification studies rely on parallel corpora, machine translation systems, or bilingual dictionaries. In \cite{cross_first, cross_second},  English labeled data are projected to Korean with parallel corpora to train relation classification models in Korean. \citet{open_cross} apply a machine translation system to translate the sentence in a target language to a source language, so that a relation classification model trained with the source language can be used. \citet{gan_cross} make use of a Generative Adversarial Network (GAN) to transfer the feature representations from the source language to the target language with the help of machine translation systems. \citet{ibm} employ bilingual word embedding mappings trained with bilingual dictionaries to develop a cross-lingual relation classification model.

In many studies, the multilingual ACE05 \cite{ace05} relation classification dataset has been treated as cross-lingual for evaluation. ACE05 includes data in English, Arabic, and Chinese; however, it is not freely available, and the number of relations is rather small, which is 6. In \cite{ibm}, a relation classification dataset for 6 languages with 53 relation types has been used, yet this dataset is not publicly available. In this paper, we release the RELX dataset created with human annotations and the RELX-Distant dataset compiled using distant supervision. Both datasets are made publicly available for future cross-lingual relation classification studies.

\section{The RELX and RELX-Distant Datasets}
\label{sec:dataset}

\begin{table}
\centering
\begin{tabular}{|l|l|c|c|}
\hline
\textbf{Dataset} & \begin{tabular}{@{}c@{}}\textbf{Total} \\ \textbf{Sentences}\end{tabular} & \begin{tabular}{@{}c@{}}\textbf{Average} \\ \textbf{Chars}\end{tabular} & \begin{tabular}{@{}c@{}}\textbf{Average} \\ \textbf{Words}\end{tabular} \\
\hline
\multicolumn{4}{|l|}{\textbf{KBP-37}}\\
\hline
Train & 15917 & 181.21 & 30.28 \\
Dev & 1724 & 181.77 & 30.55 \\
Test & 3405 & 180.20 & 30.23 \\
\hline
\multicolumn{4}{|l|}{\textbf{RELX}}\\
\hline
English & 502 & 171.18 & 28.88 \\
French & 502 & 186.63 & 30.99 \\
German & 502 & 188.27 & 27.73 \\
Spanish & 502 & 188.37 & 31.85 \\
Turkish & 502 & 170.76 & 23.60 \\

\hline
\end{tabular}
\caption{\label{relx_stat_min}
Comparative statistics of KBP-37 and RELX in different languages. Turkish translations have a lower number of words on average in the sentences due to the agglutinative nature of Turkish. The characters and words represent the average length of sentences in the corresponding dataset. 
}
\end{table}

\begin{figure*}
\centering
\begin{tabular}{l l}
\textbf{English} &  \textit{\textless e1\textgreater\space  Hoyte  \textless /e1\textgreater}\space was born in  \textit{\textless e2\textgreater\space  Guyana  \textless /e2\textgreater}\space 's capital Georgetown. \\[1mm]
\textbf{French} &  \textit{\textless e1\textgreater\space  Hoyte  \textless /e1\textgreater}\space est né à Georgetown, la capitale d' \textit{\textless e2\textgreater\space  Guyane  \textless /e2\textgreater}\space. \\[1mm]
\textbf{German} &  \begin{tabular}{@{}l@{}} \textit{\textless e1\textgreater\space  Hoyte  \textless /e1\textgreater}\space wurde in der Hauptstadt Georgetown von  \textit{\textless e2\textgreater\space  Guyana  \textless /e2\textgreater}\space \\ geboren. \end{tabular} \\[3mm]
\textbf{Spanish} &  \textit{\textless e1\textgreater\space  Hoyte  \textless /e1\textgreater}\space nació en la capital de  \textit{\textless e2\textgreater\space  Guyana  \textless /e2\textgreater}\space, Georgetown. \\[1mm]
\textbf{Turkish} &  \textit{\textless e1\textgreater\space  Hoyte  \textless /e1\textgreater}\space,  \textit{\textless e2\textgreater\space  Guyana  \textless /e2\textgreater}\space'nın başkenti Georgetown'da doğdu. \\[2mm]

\textbf{Category} & \textit{per:country\_of\_birth(e1,e2)}\\
\end{tabular}

\caption{\label{relx_sample}
Sample parallel sentences from RELX in different languages.
}
\end{figure*}

In this work, the training set of KBP-37 \cite{kbp37} is used as a source in the English language for training. For evaluation, we introduce and make publicly available the RELX dataset in English, French, German, Spanish, and Turkish. We also present RELX-Distant, which we use for the pretraining procedure in the developed MTMB (Matching the Multilingual Blanks) approach, explained in Section \ref{section:mtmb}.

\subsection{RELX}

We use the commonly-used KBP-37 English relation classification dataset for training due to its high amount of available training data. It contains 18 directional relations and a \textit{no\_relation} class, which results in \textbf{37} different classes. The statistics about KBP-37 are given in Table \ref{relx_stat_min}.

To create a cross-lingual relation classification benchmark, we selected a subset of $502$ sentences from KBP-37's test set by preserving the class distribution and the statistical features of KBP-37. 10,000 different subset selections are performed by conforming to the class distribution of KBP-37. The subset that is most similar to KBP-37 in terms of the sum of the normalized average character length and normalized average word length is chosen as the RELX dataset. Average character/word length normalization is performed by dividing to the average character/word length in the original KBP-37 test dataset. Due to the variety in the languages, the average number of characters and words in the sentences can differ for different languages, but the RELX-English and KBP-37 test set have similar distributions as summarized in Table \ref{relx_stat_min}. The average sentence length in the RELX-English dataset is slightly less than the KBP-37 test set, since we filtered problematic sentences that included URLs or consisted of more than one sentence.

The selected sentences are translated into French, German, Spanish, and Turkish by bilingual speakers who are advanced or native in both languages. They also marked the entities with (\textless e1\textgreater, \textless /e1\textgreater) and (\textless e2\textgreater, \textless /e2\textgreater) tags to match the same entities in these languages. Finally, professional translators from El Turco language services provider (\url{eltur.co}) performed language quality assessment for a randomly selected subset of RELX, containing 10\% of the sentences from each language.
Except article and synonym mistakes, there were less than three sentences with errors in each language and no critical errors were found in any of the translations. In Figure \ref{relx_sample}, we show an example of a parallel sentence from RELX with the marked entities for a sample relation.

\subsection{RELX-Distant}

\begin{table}
\centering
\begin{tabular}{|l|c|}
\hline \textbf{Language} & \textbf{Number of Sentences} \\ \hline
English & 815689 \\
French & 652842 \\
German & 652062 \\
Spanish & 397875 \\
Turkish & 57114 \\
\hline
\end{tabular}
\caption{\label{relx_distant} Number of sentences with a relation in each language in RELX-Distant.}
\end{table}

We collected a large number of multilingual sentences with relations from Wikipedia and Wikidata by a distant supervision scheme \cite{dist} and created the RELX-Distant weakly-labeled dataset for relation classification in English, French, German, Spanish, and Turkish. 

The following steps are used to create RELX-Distant:
\begin{enumerate}
    \item The Wikipedia dumps for the corresponding languages are downloaded and converted into raw documents with Wikipedia hyperlinks in entities.
    \item The raw documents are split into sentences with spaCy \cite{spacy}, and all hyperlinks, which refer to entities, are converted to their corresponding Wikidata IDs.
    \item Sentences that include entity pairs with Wikidata relations \cite{wikidata} are collected. 
\end{enumerate}

The statistics about the created \textbf{RELX-Distant} dataset are provided in Table~\ref{relx_distant}. After merging similar relations such as \textit{capital} and \textit{capital of}, RELX-Distant contains the following $24$ relations, each of which include at least 1000 sentences in English Wikipedia.\\ \textit{author}, \textit{capital}, \textit{characters}, \textit{continent}, \textit{country of citizenship}, \textit{country of origin}, \textit{developer}, \textit{ethnic group}, \textit{father}, \textit{instance of}, \textit{language}, \textit{located in country}, \textit{member of}, \textit{mother}, \textit{owned by}, \textit{parent organization}, \textit{parent taxon}, \textit{part of}, \textit{partner}, \textit{performer}, \textit{place of birth}, \textit{religion}, \textit{sibling}, \textit{spouse}

\section{Methods}

\begin{figure*}
    \centering
    \includegraphics[height=6cm,keepaspectratio]{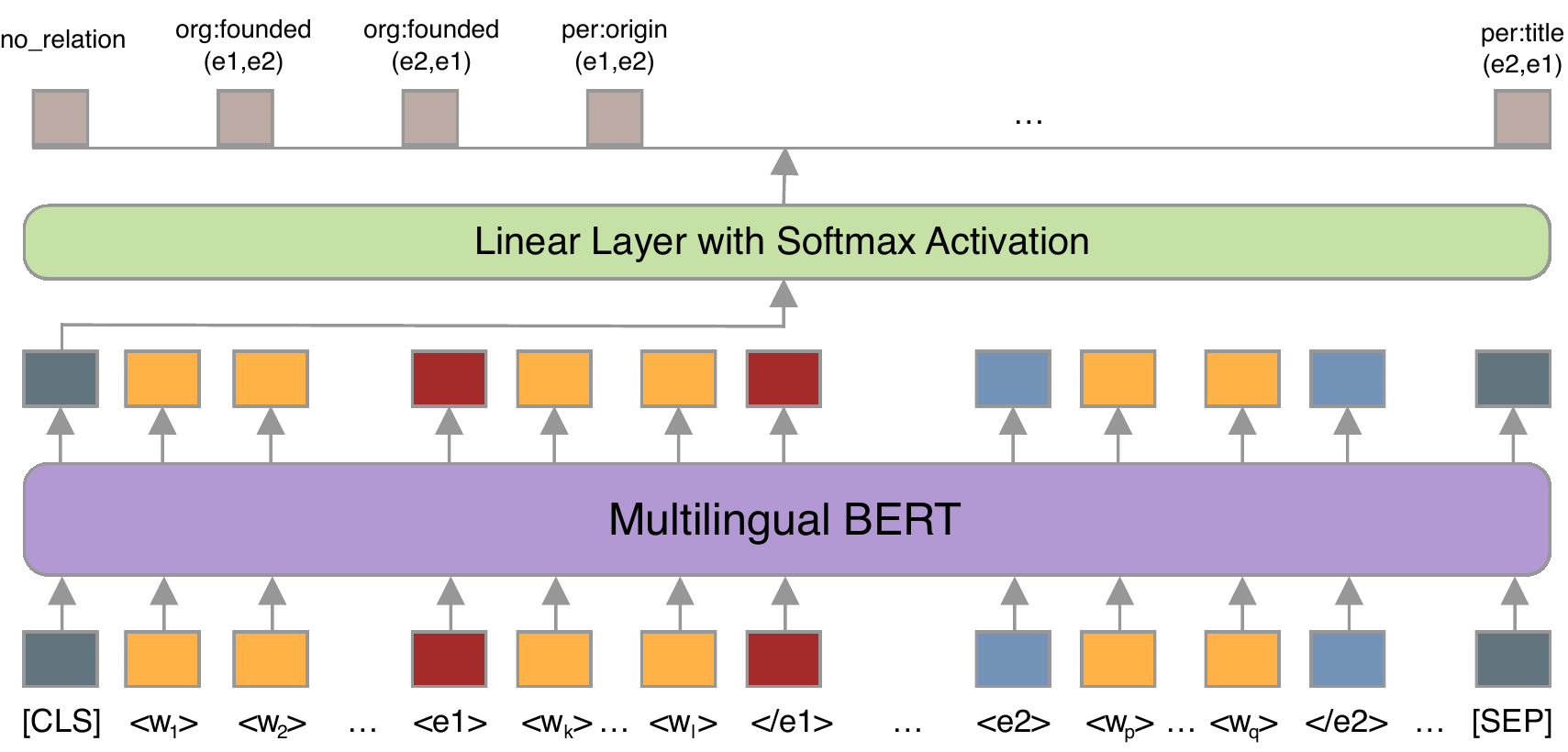}
    \caption{Illustration of our model. \textless w\textsubscript{i}\textgreater 's represent tokens from BERT tokenizer, \textless e1\textgreater,  \textless /e1\textgreater\space and \textless e2\textgreater,  \textless /e2\textgreater\space represent entity start and end markers for the first and second entities, respectively. [CLS] and [SEP] are special tokens in BERT. [CLS] can be used as a fixed-length input representation and [SEP] denotes the end of the sentence.}
    \label{fig:model}
\end{figure*}

\label{sec:methods}
\textbf{Task Definition:} In the cross-lingual relation classification task, we are given a source language dataset $D_s$ with $n_s$ sentences containing related entity pairs.
\begin{align*}
    D_s & = \{(S^{s}_{i}, E1^{s}_{i}, E2^{s}_{i}, r_{i})\}_{i=1}^{i=n_s} \text{  where}\\
    S^{s}_i & = [w_1, w_2, ..., w_n] \\
    E1^{s}_i & = (w_k, w_{k+1}, ..., w_l) \\
    E2^{s}_i & = (w_p,w_{p+1}, ..., w_q) \\
    r_i &\in R
\end{align*}

$E1^{s}_{i}$ and $E2^{s}_{i}$ correspond to entities and $w_i$ correspond to tokens in the sentence $S^{s}_{i}$. $r_{i}$ is the directional relation between $E1^{s}_{i}$ and $E2^{s}_{i}$ in $S^{s}_{i}$, selected from a predefined relation set $R$.

Given test set $D_t = \{(S^{t}_{j}, E1^{t}_{j}, E2^{t}_{j})\}_{j=1}^{j=n_t}$ in the target language, cross-lingual relation classification aims to find the relation probability $P(r_j|S^{t}_j, E1^{t}_j, E2^{t}_j)$ where $r_j \in R$ for a sentence and an entity pair in the target language with the supervision of $D_s$ in the source language.

\subsection{Multilingual BERT}\label{mbert}

Multilingual BERT (mBERT) \cite{bert}, is a multilingual language model trained on 104 languages using the corresponding Wikipedia dumps. Due to shared word pieces like URLs and numbers across languages \cite{word_order}, mBERT is able to produce fixed-length sentence representations for these languages. Exponential smoothed weighting is used in order to reduce the under-representation problem of low-resource languages that have a relatively smaller number of Wikipedia articles.

mBERT is selected as our baseline model in this work, similar to recent cross-lingual tasks such as natural language inference \cite{bad_turkish} and question answering \cite{xquad}. Each sentence is tokenized by the mBERT tokenizer. Following \cite{mtb}, entity markers are added to emphasize the locations of the entities in the sentences. We add entity start and end markers that are special tokens, which are learned from scratch during the training, as shown in Figure \ref{fig:model}.

\begin{figure*}
\centering
\begin{tabular}{|c|l|}
 \multicolumn{2}{l}{\textbf{Sentences}} \\
\hline
S\textsubscript{en} & \begin{tabular}[t]{@{}l}
    In the 3rd century, \textbf{E2} wrote his ``\textbf{E1}'' and other exegetical and theological\\ works while living in Caesarea.
\end{tabular}\\[6mm]
S\textsubscript{es} & 
\begin{tabular}[t]{@{}l}
Este es un palimpsesto de una copia de la obra de \textbf{E2} llamada la \textbf{E1}.\footnotemark
\end{tabular}\\[2mm]
S\textsubscript{tr} &  
\begin{tabular}[t]{@{}l}
İreneyus ve \textbf{E2} gibi kilise babalarının metinlerinde aktarılanlara göre esasen\\ \textbf{E3}li olan Marcellina, Anicetus döneminde Roma'ya göç etmiş ve çok\\ sayıda takipçi toplamıştır.\footnotemark
\end{tabular}\\

\hline
 \multicolumn{2}{l}{\textbf{Entities}} \\
\hline 
    E1 & \textit{Q839739} (Hexapla, Hexapla, Hexapla) \\
    E2 & \textit{Q170472} (Origen, Orígenes, Origenes)\\
    E3 & \textit{Q87} (Alexandria, Alejandría, İskenderiye)\\
\hline
 \multicolumn{2}{l}{\textbf{Relations}} \\
\hline
 (E1, E2) & \textit{P50} (Author) \\
 (E2, E3) & \textit{P19} (Place of Birth) \\
\hline
 \multicolumn{2}{l}{\textbf{Pairs}} \\
\hline
    Positive& (S\textsubscript{en}, S\textsubscript{es}) \\
    Negative&(S\textsubscript{en}, S\textsubscript{tr}) \\
\hline
\end{tabular}

\caption{\label{mtmb_data_fig}
Sample positive and negative pairs constructed from RELX-Distant. Entities and relations are linked with their Wikidata ID's (shown in italic) and words in parentheses in entities represent English, Spanish, and Turkish correspondence, consecutively. }
\end{figure*}

Our objective is to predict the relation between a given entity pair in a sentence from among a set of relations. 
For this purpose, as in  \cite{bert}, mBERT's output state of the [CLS] token is used as fixed-length sentence representation (or in our case as relation representation). This representation is fed into a linear layer with softmax activation to predict the probability of each relation, as illustrated in Figure \ref{fig:model}. The developed model predicts the probabilities of the \textit{no\_relation} class and 18 directional relation classes, which result in 37 different classes in the KBP-37 and RELX datasets.

Our implementation details about mBERT are as follows.
\begin{itemize}
    \item We use the initial weights of Cased Multilingual BERT from \cite{bert}, which has 12 layers, 768 hidden size, 12 heads, and 110M parameters.
    \item The network on top of the transformer architecture that gets the [CLS] representation as input for relation class prediction has a linear layer with softmax activation. 
    \item AdamW with $3e-5$ learning rate and $0.1$ weight decay is used with a batch size of $64$.
    \item The classification loss is selected as the cross-entropy of the predictions with respect to the true labels.  
\end{itemize}

\footnotetext[3]{\textit{English Translation:} This is a palimpsest of a copy of \textbf{E2}'s work called \textbf{E1}.}
\footnotetext{\textit{English Translation:} According to what was reported in the texts of the church fathers such as Irenaeus and \textbf{E2}, Marcellina, who was originally from \textbf{E3}, migrated to Rome during the Anicetus period and collected many followers.}

The hyperparameters are tuned over the KBP-37 validation set based on the F1-score as described in Section \ref{sec:results}. Learning rates of 1e-3, 1e-4, 3e-4, 1e-5, 3e-5, and 1e-6; weight decay values of 0, 0.01, and 0.1 with the SGD, Adam \cite{adam}, and AdamW \cite{adamw} optimizers have been evaluated with PyTorch \cite{pytorch} and HuggingFace's Transformers \cite{huggingface} libraries. The best values have been determined as $3e-5$ learning rate and $0.1$ weight decay with the  AdamW optimizer.

\subsection{Matching the Multilingual Blanks}
\label{section:mtmb}

Our objective is to pretrain a public checkpoint of mBERT, released by \cite{bert}, in a way that it can learn various representations of relations across different languages. In order to do this, we prepare RELX-Distant, whose entities are labeled by using Wikipedia hyperlinks, to create pairs of sentences from different languages and propose Matching the Multilingual Blanks, a multilingual distant supervision approach that targets detecting the similarity between the relations described in an input multilingual pair of sentences.

For this model, we pretrain mBERT with two objectives: Masked Language Model from \cite{bert} and Matching the Multilingual Blanks (MTMB). Similar to the monolingual work in \cite{mtb}, we create positive and negative multilingual sentence pairs from RELX-Distant for the MTMB objective. We pretrain mBERT with the aim of learning how relations are represented in different languages by predicting whether the English sentence and the non-English sentence in a pair have the same relation or not.

Positive sentence pairs are selected to share the same entities, which result in having the same relation by the distant supervision scheme. $(S_{en}, S_{es})$ in Figure \ref{mtmb_data_fig} is a positive pair because both sentences include the E1 (Hexapla) and E2 (Origen) entities that have the P50 (Author) relation. 

In the negative sentence pairs, each sentence has entities with different relations. In order to avoid dissimilar sentences in a negative pair, which may cause our model to make predictions based on the topics of the sentences, we use \textit{strong} negative pairs similar to \cite{mtb}. In strong negative pairs, one of the entities in each sentence in the pair is common. $(S_{en}, S_{tr})$ in Figure \ref{mtmb_data_fig} is a strong negative pair because both sentences share the entity E2 (Origen), but the English sentence has the P50 (Author) relation, and the Turkish sentence has the P19 (Place of Birth) relation.

In the compiled sentence pairs, the entities are replaced by a special [BLANK] token with $0.7$ probability to capture the text patterns better and avoid memorizing the entities. By following these steps, we create 20 million pairs of sentences from RELX-Distant to pretrain mBERT. These sentence pairs have a uniform distribution with respect to the positive and negative classes as well as the languages in RELX-Distant. 
We call the pretraining procedure of mBERT with multilingual sentence pairs, Matching the Multilingual Blanks (MTMB).

The implementation details of the model are similar to the model described in Section \ref{mbert}. However, before multi-way relation classification training, we first pretrain the public checkpoint of mBERT \cite{bert} with two objectives. The first objective is the Masked Language Model, and we implement it as implemented in \cite{bert}. The second objective is a binary classification of sentence pairs, whether two sentences in different languages have the same relation or not. While fine-tuning mBERT in Section \ref{mbert} is relatively inexpensive (less than 10 minutes in each epoch on a GPU), one epoch of MTMB with 20 millions of sentence pairs takes approximately 10 days on a Tesla V100 GPU. Considering this, we release the weights of our MTMB model publicly in \url{https://github.com/boun-tabi/RELX}.

\section{Results}
\label{sec:results}

We compare our monolingual relation classification results using KBP-37 and the cross-lingual results using RELX. We report our results by taking the average scores of 10 runs to decrease the effect of high variance between different runs in BERT as stated in \cite{bert_variance}.\\
\textbf{Evaluation Metric:} We use (18+1)-way evaluation by taking directionality into account as used in \cite{semeval}. First, the F1 score of a relation is calculated by taking the micro average of F1's of both directions. Then, the macro average of F1 scores of 18 relations is considered as our final score. 
\subsection{KBP-37}

\begin{table}
\centering
\begin{tabular}{|l|c|c|}
\hline \textbf{Model} & \textbf{Dev} & \textbf{Test} \\ \hline
BERT\textsubscript{Large} \cite{mtb}& 69.5 & 68.3\\
MTB\quad\quad\space\space\space \cite{mtb}& \textbf{70.3} & \textbf{69.3}\\
\hline
BERT\textsubscript{Base} & 66.0  & 65.4 \\
mBERT & 65.5  & 64.9 \\
MTMB & \textbf{66.8} & \textbf{66.5} \\
\hline
\end{tabular}
\caption{\label{kbp37_res} F1 scores of our models compared to the state-of-the-art models on the development and test sets of \textbf{KBP-37} (\textit{English}).}
\end{table}

Table \ref{kbp37_res} contains the results for our models and the state-of-the-art models evaluated on the KBP-37 development and test sets. BERT\textsubscript{Large} and MTB are models from \cite{mtb}. Both models use pretrained BERT\textsubscript{Large}, which is specific to the English language. 
We finetune three models for relation classification with the same architecture and number of parameters: BERT\textsubscript{Base}, mBERT, and MTMB; where mBERT and MTMB are pretrained on multilingual corpora, while BERT\textsubscript{Base} is pretrained on English corpora.
The complexity of BERT\textsubscript{Large} is much higher than mBERT and BERT\textsubscript{Base}. The number of parameters in BERT\textsubscript{Large} is 340 million, while mBERT and BERT\textsubscript{Base} have 110 million parameters. Also, BERT\textsubscript{Large} has 24 layers and 16 heads compared to 12 layers and 12 heads in mBERT and BERT\textsubscript{Base}. Finally, the hidden size in BERT\textsubscript{Large} is 1024, while it is 768 in mBERT and BERT\textsubscript{Base}. Because of the difference in complexity and the language of the training data, as expected, BERT\textsubscript{Large} based models have better performance for the English language than mBERT based models. Still, the results show that Matching the Multilingual Blanks significantly ($\text{p-value}<0.05$) outperforms mBERT and BERT\textsubscript{Base} in the English language according to the randomization tests \cite{randomization}.

\subsection{RELX}

\begin{figure*}
    \centering
    \includegraphics[height=10cm,keepaspectratio]{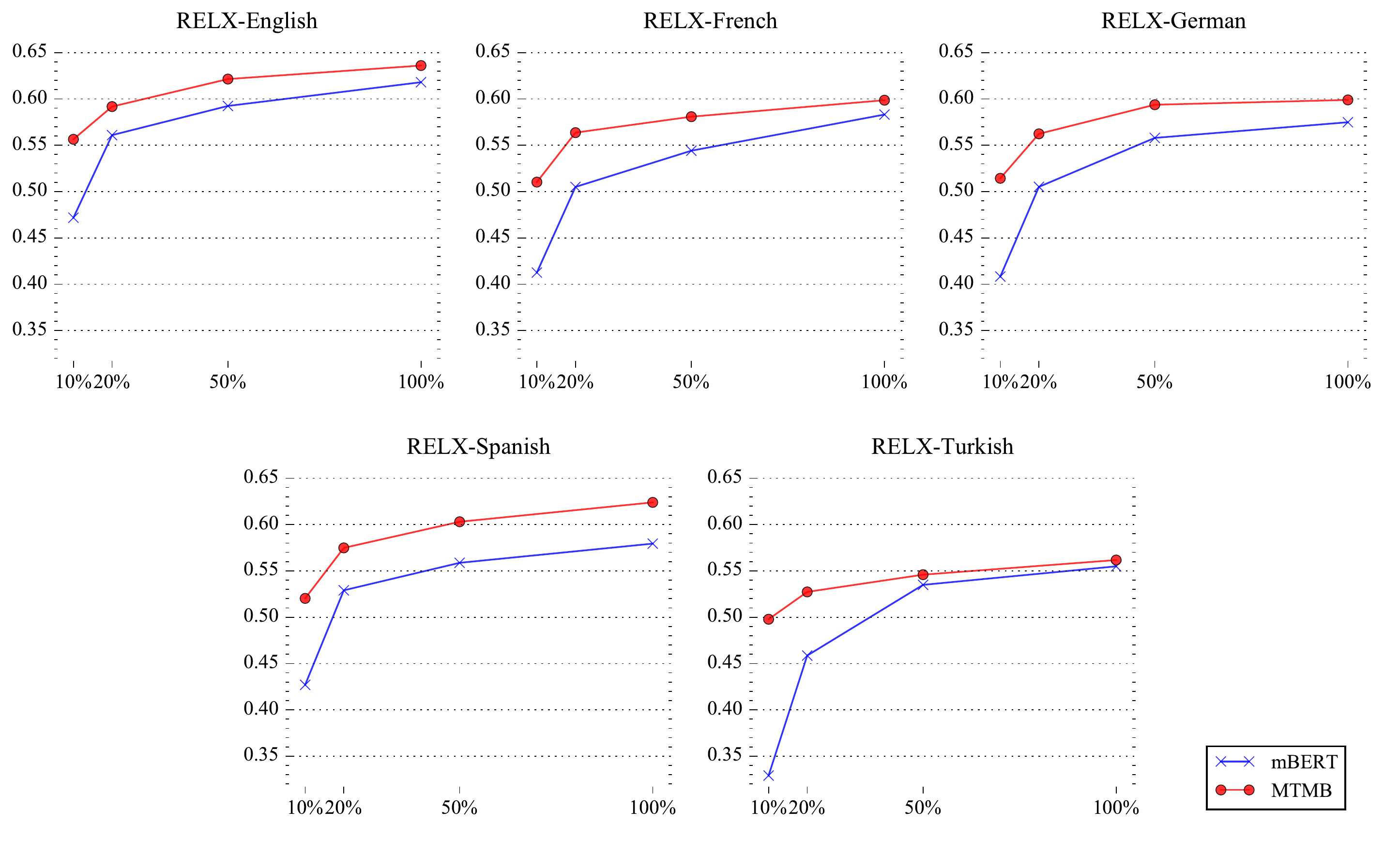}
    \caption{\label{fig:varied_results} Cross-lingual relation classification performance (F1 score \textit{y-axis}) of mBERT and MTMB with varying amounts of training data (\textit{x-axis}). }
\end{figure*}

\begin{table}
\centering
\begin{tabular}{|l|c|c|c|c|c|}
\hline \textbf{Model} & \textbf{EN} & \textbf{FR} & \textbf{DE} & \textbf{ES} & \textbf{TR} \\ \hline
mBERT & 61.8  & 58.3 & 57.5 & 57.9 & 55.8 \\
MTMB & \textbf{63.6} & \textbf{59.9} & \textbf{59.9} & \textbf{62.4} & \textbf{56.2}  \\
\hline
\end{tabular}
\caption{\label{relx_res} F1 scores of mBERT and MTMB evaluated on RELX. The columns represent the English, French, German, Spanish, and Turkish parts of \textbf{RELX}.}
\end{table}

RELX is used to evaluate the mBERT and MTMB models, which are finetuned on the training set (which is in English) of KBP-37. The results are summarized in Table \ref{relx_res}. By Matching the Multilingual Blanks training setup, we significantly ($\text{p-value}<0.05$) improve the performance of mBERT for five languages, including RELX-English. In cross-lingual cases and the monolingual case, MTMB significantly outperforms mBERT based on the randomization tests. 

We display the results by varying the size of the training data in Figure \ref{fig:varied_results}. The results show that MTMB performs better than mBERT, especially in low-resource cases. The difference in F1 scores between MTMB and mBERT is more significant when the amount of the available training data is lower. For Spanish, MTMB was able to reach the performance of mBERT that uses all the training data by using only around 20\% of the training data, and for the other evaluated languages (except Turkish), around 50\% of the data was sufficient to obtain the same performance as mBERT that uses all the training data. Thus, the required human annotations in the source language can be significantly reduced with the help of MTMB.

Table \ref{relx_res} demonstrates that the best cross-lingual performance is achieved for Spanish, which is on par with prior studies on other cross-lingual NLP tasks such as question answering and natural language inference \cite{xquad} that also report higher performance for Spanish. On the other hand, our results show that the worse cross-lingual performance is obtained for Turkish. 
\citet{word_order} observe that mBERT performance is effected by word ordering and works best for typologically similar languages. In order to investigate this, we compare the source language (English) and target languages (French, German, Spanish, Turkish) by a subset of the World Atlas of Language Structures (WALS) features \cite{wals} that are relevant to grammatical ordering\footnote{\textbf{81A}: Order of Subject, Object and Verb, \textbf{85A}: Order of Adposition and Noun Phrase, \textbf{86A}: Order of Genitive and Noun, \textbf{87A}: Order of Adjective and Noun, \textbf{88A}: Order of Demonstrative and Noun, \textbf{89A}: Order of Numeral and Noun} as in \cite{word_order}. Considering these features, Turkish is the least similar language to English among the
languages in RELX. Our results support the claim presented in \cite{word_order}.

Error analysis reveals that 120 out of 176 mispredicted sentences in RELX-English are common in all target languages. Among these common errors, classes with less than 600 samples in the training data have 60\% more error rate, suggesting that increasing their number of samples may benefit in all languages.

We also analyzed relation direction errors, where the predicted relation class is the same as the gold class, while the predicted direction is incorrect. There are 79 relation direction errors for Turkish, whereas there are less than 15 for the other languages. Turkish has generally an SOV word order and postpositions, while English has generally SVO word order and prepositions. These differences between Turkish and English are possible causes for the problems related to direction errors as discussed in \cite{word_order}. Finally, no notable difference is observed in errors across languages in terms of sentence length.

\section{Conclusion}
\label{sec:conclusion}
In this paper, we addressed the cross-lingual relation classification task. First, we introduced two publicly available datasets: \textbf{RELX}, a cross-lingual relation classification benchmark for English, French, German, Spanish, and Turkish with parallel sentences and \textbf{RELX-Distant}, a multilingual dataset containing a large number of sentences with relations from Wikipedia and Wikidata collected via distant supervision. Second, we proposed a baseline model with mBERT and a new multilingual pretraining scheme with distant supervision called Matching the Multilingual Blanks (MTMB). Our experiments showed that MTMB significantly outperforms the mBERT baseline on the monolingual and cross-lingual datasets. The improvement obtained by MTMB is higher in the low-resource settings for the source language.
We also showed that better cross-lingual relation classification performance is obtained for target languages which are typologically similar to the source language. The performance for Spanish is comparable to  English (the source language in this study), while the lowest F1 scores are obtained for Turkish. MTMB can be easily adopted to other languages by using our provided scripts\footnote{\hbox{\url{https://github.com/boun-tabi/RELX}}}. The only requirement is the availability of Wikipedia articles in the new target language. 

As future work, we plan to extend RELX-Distant to all the available languages in Wikipedia. We will also investigate the effect of MTMB in different cross-lingual tasks such as natural language inference, named entity recognition, and question answering by using the extended RELX-Distant dataset.

\section*{Acknowledgments}
We would like to thank Ahmed Yusuf Asilt\"urk, Barbaros Eri\c{s}, Esmanur Y{\i}lmaz, Gizem \"Ozkanal, Ramazan Pala, and Taha K\"uç\"ukkat{\i}rc{\i} for their contributions on the translation and \url{eltur.co} for their quality check service. TUBITAK-BIDEB 2211-A Scholarship Program (to A.K.), and TUBA-GEBIP Award of the Turkish Science Academy (to A.O.) are gratefully acknowledged.

\bibliographystyle{acl_natbib}
\bibliography{emnlp2020}

\begin{thebibliography}{40}
\expandafter\ifx\csname natexlab\endcsname\relax\def\natexlab#1{#1}\fi

\bibitem[{Abdou et~al.(2019)Abdou, Sas, Aralikatte, Augenstein, and
  S{\o}gaard}]{abdou2019x}
Mostafa Abdou, Cezar Sas, Rahul Aralikatte, Isabelle Augenstein, and Anders
  S{\o}gaard. 2019.
\newblock {X-WikiRE}: A large, multilingual resource for relation extraction as
  machine comprehension.
\newblock In \emph{Proceedings of the 2nd Workshop on Deep Learning Approaches
  for Low-Resource NLP (DeepLo 2019)}, pages 265--274.

\bibitem[{Artetxe et~al.(2020)Artetxe, Ruder, and Yogatama}]{xquad}
Mikel Artetxe, Sebastian Ruder, and Dani Yogatama. 2020.
\newblock \href {https://doi.org/10.18653/v1/2020.acl-main.421} {On the
  cross-lingual transferability of monolingual representations}.
\newblock In \emph{Proceedings of the 58th Annual Meeting of the Association
  for Computational Linguistics}, pages 4623--4637, Online. Association for
  Computational Linguistics.

\bibitem[{Conneau et~al.(2020)Conneau, Khandelwal, Goyal, Chaudhary, Wenzek,
  Guzm{\'a}n, Grave, Ott, Zettlemoyer, and Stoyanov}]{xlm_roberta}
Alexis Conneau, Kartikay Khandelwal, Naman Goyal, Vishrav Chaudhary, Guillaume
  Wenzek, Francisco Guzm{\'a}n, Edouard Grave, Myle Ott, Luke Zettlemoyer, and
  Veselin Stoyanov. 2020.
\newblock \href {https://doi.org/10.18653/v1/2020.acl-main.747} {Unsupervised
  cross-lingual representation learning at scale}.
\newblock In \emph{Proceedings of the 58th Annual Meeting of the Association
  for Computational Linguistics}, pages 8440--8451, Online. Association for
  Computational Linguistics.

\bibitem[{Conneau and Lample(2019)}]{xlm}
Alexis Conneau and Guillaume Lample. 2019.
\newblock Cross-lingual language model pretraining.
\newblock In \emph{Advances in Neural Information Processing Systems}, pages
  7059--7069.

\bibitem[{Conneau et~al.(2018)Conneau, Rinott, Lample, Williams, Bowman,
  Schwenk, and Stoyanov}]{nli_wemb}
Alexis Conneau, Ruty Rinott, Guillaume Lample, Adina Williams, Samuel Bowman,
  Holger Schwenk, and Veselin Stoyanov. 2018.
\newblock \href {https://doi.org/10.18653/v1/D18-1269} {{XNLI}: Evaluating
  cross-lingual sentence representations}.
\newblock In \emph{Proceedings of the 2018 Conference on Empirical Methods in
  Natural Language Processing}, pages 2475--2485, Brussels, Belgium.
  Association for Computational Linguistics.

\bibitem[{Devlin et~al.(2019)Devlin, Chang, Lee, and Toutanova}]{bert}
Jacob Devlin, Ming-Wei Chang, Kenton Lee, and Kristina Toutanova. 2019.
\newblock \href {https://doi.org/10.18653/v1/N19-1423} {{BERT}: Pre-training of
  deep bidirectional transformers for language understanding}.
\newblock In \emph{Proceedings of the 2019 Conference of the North {A}merican
  Chapter of the Association for Computational Linguistics: Human Language
  Technologies, Volume 1 (Long and Short Papers)}, pages 4171--4186,
  Minneapolis, Minnesota. Association for Computational Linguistics.

\bibitem[{Dodge et~al.(2020)Dodge, Ilharco, Schwartz, Farhadi, Hajishirzi, and
  Smith}]{bert_variance}
Jesse Dodge, Gabriel Ilharco, Roy Schwartz, Ali Farhadi, Hannaneh Hajishirzi,
  and Noah Smith. 2020.
\newblock Fine-tuning pretrained language models: Weight initializations, data
  orders, and early stopping.
\newblock \emph{arXiv preprint arXiv:2002.06305}.

\bibitem[{Dryer and Haspelmath(2013)}]{wals}
Matthew~S. Dryer and Martin Haspelmath, editors. 2013.
\newblock \href {https://wals.info/} {\emph{WALS Online}}.
\newblock Max Planck Institute for Evolutionary Anthropology, Leipzig.

\bibitem[{Faruqui and Kumar(2015)}]{open_cross}
Manaal Faruqui and Shankar Kumar. 2015.
\newblock \href {https://doi.org/10.3115/v1/N15-1151} {Multilingual open
  relation extraction using cross-lingual projection}.
\newblock In \emph{Proceedings of the 2015 Conference of the North {A}merican
  Chapter of the Association for Computational Linguistics: Human Language
  Technologies}, pages 1351--1356, Denver, Colorado. Association for
  Computational Linguistics.

\bibitem[{Hendrickx et~al.(2010)Hendrickx, Kim, Kozareva, Nakov,
  {\'O}~S{\'e}aghdha, Pad{\'o}, Pennacchiotti, Romano, and
  Szpakowicz}]{semeval}
Iris Hendrickx, Su~Nam Kim, Zornitsa Kozareva, Preslav Nakov, Diarmuid
  {\'O}~S{\'e}aghdha, Sebastian Pad{\'o}, Marco Pennacchiotti, Lorenza Romano,
  and Stan Szpakowicz. 2010.
\newblock \href {https://www.aclweb.org/anthology/S10-1006} {{S}em{E}val-2010
  task 8: Multi-way classification of semantic relations between pairs of
  nominals}.
\newblock In \emph{Proceedings of the 5th International Workshop on Semantic
  Evaluation}, pages 33--38, Uppsala, Sweden. Association for Computational
  Linguistics.

\bibitem[{Honnibal and Montani(2017)}]{spacy}
Matthew Honnibal and Ines Montani. 2017.
\newblock {spaCy 2}: Natural language understanding with {B}loom embeddings,
  convolutional neural networks and incremental parsing.
\newblock To appear.

\bibitem[{Indurkhya(2015)}]{text_mining}
Nitin Indurkhya. 2015.
\newblock \href {https://doi.org/10.1002/widm.1154} {Emerging directions in
  predictive text mining}.
\newblock \emph{WIREs Data Mining and Knowledge Discovery}, 5(4):155--164.

\bibitem[{Kambhatla(2004)}]{handcraft1}
Nanda Kambhatla. 2004.
\newblock \href {https://www.aclweb.org/anthology/P04-3022} {Combining lexical,
  syntactic, and semantic features with maximum entropy models for information
  extraction}.
\newblock In \emph{Proceedings of the {ACL} Interactive Poster and
  Demonstration Sessions}, pages 178--181, Barcelona, Spain. Association for
  Computational Linguistics.

\bibitem[{Kim et~al.(2010)Kim, Jeong, Lee, and Lee}]{cross_first}
Seokhwan Kim, Minwoo Jeong, Jonghoon Lee, and Gary~Geunbae Lee. 2010.
\newblock \href {https://www.aclweb.org/anthology/C10-1064} {A cross-lingual
  annotation projection approach for relation detection}.
\newblock In \emph{Proceedings of the 23rd International Conference on
  Computational Linguistics (Coling 2010)}, pages 564--571, Beijing, China.
  Coling 2010 Organizing Committee.

\bibitem[{Kim and Lee(2012)}]{cross_second}
Seokhwan Kim and Gary~Geunbae Lee. 2012.
\newblock \href {https://www.aclweb.org/anthology/P12-2010} {A graph-based
  cross-lingual projection approach for weakly supervised relation extraction}.
\newblock In \emph{Proceedings of the 50th Annual Meeting of the Association
  for Computational Linguistics (Volume 2: Short Papers)}, pages 48--53, Jeju
  Island, Korea. Association for Computational Linguistics.

\bibitem[{Kingma and Ba(2015)}]{adam}
Diederik~P. Kingma and Jimmy Ba. 2015.
\newblock \href {http://arxiv.org/abs/1412.6980} {Adam: {A} method for
  stochastic optimization}.
\newblock In \emph{3rd International Conference on Learning Representations,
  {ICLR} 2015, San Diego, CA, USA, May 7-9, 2015, Conference Track
  Proceedings}.

\bibitem[{Liu et~al.(2019)Liu, Lin, Liu, and Sun}]{xqa}
Jiahua Liu, Yankai Lin, Zhiyuan Liu, and Maosong Sun. 2019.
\newblock \href {https://doi.org/10.18653/v1/P19-1227} {{XQA}: A cross-lingual
  open-domain question answering dataset}.
\newblock In \emph{Proceedings of the 57th Annual Meeting of the Association
  for Computational Linguistics}, pages 2358--2368, Florence, Italy.
  Association for Computational Linguistics.

\bibitem[{Loshchilov and Hutter(2018)}]{adamw}
Ilya Loshchilov and Frank Hutter. 2018.
\newblock Decoupled weight decay regularization.
\newblock In \emph{International Conference on Learning Representations}.

\bibitem[{Mikolov et~al.(2013)Mikolov, Sutskever, Chen, Corrado, and
  Dean}]{word2vec}
Tomas Mikolov, Ilya Sutskever, Kai Chen, Greg~S Corrado, and Jeff Dean. 2013.
\newblock Distributed representations of words and phrases and their
  compositionality.
\newblock In \emph{Advances in Neural Information Processing Systems}, pages
  3111--3119.

\bibitem[{Mintz et~al.(2009)Mintz, Bills, Snow, and Jurafsky}]{dist}
Mike Mintz, Steven Bills, Rion Snow, and Daniel Jurafsky. 2009.
\newblock \href {https://www.aclweb.org/anthology/P09-1113} {Distant
  supervision for relation extraction without labeled data}.
\newblock In \emph{Proceedings of the Joint Conference of the 47th Annual
  Meeting of the {ACL} and the 4th International Joint Conference on Natural
  Language Processing of the {AFNLP}}, pages 1003--1011, Suntec, Singapore.
  Association for Computational Linguistics.

\bibitem[{Nguyen and Grishman(2015)}]{cnn2}
Thien~Huu Nguyen and Ralph Grishman. 2015.
\newblock \href {https://doi.org/10.3115/v1/W15-1506} {Relation extraction:
  Perspective from convolutional neural networks}.
\newblock In \emph{Proceedings of the 1st Workshop on Vector Space Modeling for
  Natural Language Processing}, pages 39--48, Denver, Colorado. Association for
  Computational Linguistics.

\bibitem[{Ni and Florian(2019)}]{ibm}
Jian Ni and Radu Florian. 2019.
\newblock \href {https://doi.org/10.18653/v1/D19-1038} {Neural cross-lingual
  relation extraction based on bilingual word embedding mapping}.
\newblock In \emph{Proceedings of the 2019 Conference on Empirical Methods in
  Natural Language Processing and the 9th International Joint Conference on
  Natural Language Processing (EMNLP-IJCNLP)}, pages 399--409, Hong Kong,
  China. Association for Computational Linguistics.

\bibitem[{Nothman et~al.(2013)Nothman, Ringland, Radford, Murphy, and
  Curran}]{mner}
Joel Nothman, Nicky Ringland, Will Radford, Tara Murphy, and James~R Curran.
  2013.
\newblock Learning multilingual named entity recognition from {Wikipedia}.
\newblock \emph{Artificial Intelligence}, 194:151--175.

\bibitem[{Paszke et~al.(2019)Paszke, Gross, Massa, Lerer, Bradbury, Chanan,
  Killeen, Lin, Gimelshein, Antiga et~al.}]{pytorch}
Adam Paszke, Sam Gross, Francisco Massa, Adam Lerer, James Bradbury, Gregory
  Chanan, Trevor Killeen, Zeming Lin, Natalia Gimelshein, Luca Antiga, et~al.
  2019.
\newblock Pytorch: An imperative style, high-performance deep learning library.
\newblock In \emph{Advances in Neural Information Processing Systems}, pages
  8026--8037.

\bibitem[{Pennington et~al.(2014)Pennington, Socher, and Manning}]{glove}
Jeffrey Pennington, Richard Socher, and Christopher Manning. 2014.
\newblock \href {https://doi.org/10.3115/v1/D14-1162} {{G}love: Global vectors
  for word representation}.
\newblock In \emph{Proceedings of the 2014 Conference on Empirical Methods in
  Natural Language Processing ({EMNLP})}, pages 1532--1543, Doha, Qatar.
  Association for Computational Linguistics.

\bibitem[{Peters et~al.(2018)Peters, Neumann, Iyyer, Gardner, Clark, Lee, and
  Zettlemoyer}]{elmo}
Matthew Peters, Mark Neumann, Mohit Iyyer, Matt Gardner, Christopher Clark,
  Kenton Lee, and Luke Zettlemoyer. 2018.
\newblock \href {https://doi.org/10.18653/v1/N18-1202} {Deep contextualized
  word representations}.
\newblock In \emph{Proceedings of the 2018 Conference of the North {A}merican
  Chapter of the Association for Computational Linguistics: Human Language
  Technologies, Volume 1 (Long Papers)}, pages 2227--2237, New Orleans,
  Louisiana. Association for Computational Linguistics.

\bibitem[{Pires et~al.(2019)Pires, Schlinger, and Garrette}]{word_order}
Telmo Pires, Eva Schlinger, and Dan Garrette. 2019.
\newblock \href {https://doi.org/10.18653/v1/P19-1493} {How multilingual is
  multilingual {BERT}?}
\newblock In \emph{Proceedings of the 57th Annual Meeting of the Association
  for Computational Linguistics}, pages 4996--5001, Florence, Italy.
  Association for Computational Linguistics.

\bibitem[{Soares et~al.(2019)Soares, FitzGerald, Ling, and Kwiatkowski}]{mtb}
Livio~Baldini Soares, Nicholas FitzGerald, Jeffrey Ling, and Tom Kwiatkowski.
  2019.
\newblock Matching the blanks: Distributional similarity for relation learning.
\newblock In \emph{Proceedings of the 57th Annual Meeting of the Association
  for Computational Linguistics}, pages 2895--2905.

\bibitem[{Upadhyay(2019)}]{web_ratio}
Shyam Upadhyay. 2019.
\newblock \emph{Exploiting Cross-lingual Representations for Natural Language
  Processing}.
\newblock Ph.D. thesis, University of Pennsylvania.

\bibitem[{Vrande{\v{c}}i{\'c} and Kr{\"o}tzsch(2014)}]{wikidata}
Denny Vrande{\v{c}}i{\'c} and Markus Kr{\"o}tzsch. 2014.
\newblock Wikidata: a free collaborative knowledgebase.
\newblock \emph{Communications of the ACM}, 57(10):78--85.

\bibitem[{Walker et~al.(2006)Walker, Strassel, Medero, and Maeda}]{ace05}
Christopher Walker, Stephanie Strassel, Julie Medero, and Kazuaki Maeda. 2006.
\newblock \href {https://catalog.ldc.upenn.edu/LDC2006T06} {{ACE} 2005
  multilingual training corpus}.
\newblock Philadelphia: Linguistic Data Consortium.

\bibitem[{Wolf et~al.(2019)Wolf, Debut, Sanh, Chaumond, Delangue, Moi, Cistac,
  Rault, Louf, Funtowicz, Davison, Shleifer, von Platen, Ma, Jernite, Plu, Xu,
  Scao, Gugger, Drame, Lhoest, and Rush}]{huggingface}
Thomas Wolf, Lysandre Debut, Victor Sanh, Julien Chaumond, Clement Delangue,
  Anthony Moi, Pierric Cistac, Tim Rault, Rémi Louf, Morgan Funtowicz, Joe
  Davison, Sam Shleifer, Patrick von Platen, Clara Ma, Yacine Jernite, Julien
  Plu, Canwen Xu, Teven~Le Scao, Sylvain Gugger, Mariama Drame, Quentin Lhoest,
  and Alexander~M. Rush. 2019.
\newblock Huggingface's transformers: State-of-the-art natural language
  processing.
\newblock \emph{ArXiv}, abs/1910.03771.

\bibitem[{Wu and Dredze(2019)}]{bad_turkish}
Shijie Wu and Mark Dredze. 2019.
\newblock Beto, bentz, becas: The surprising cross-lingual effectiveness of
  {BERT}.
\newblock In \emph{Proceedings of the 2019 Conference on Empirical Methods in
  Natural Language Processing and the 9th International Joint Conference on
  Natural Language Processing (EMNLP-IJCNLP)}, pages 833--844.

\bibitem[{Xie et~al.(2018)Xie, Yang, Neubig, Smith, and Carbonell}]{ner_wemb}
Jiateng Xie, Zhilin Yang, Graham Neubig, Noah~A. Smith, and Jaime Carbonell.
  2018.
\newblock \href {https://doi.org/10.18653/v1/D18-1034} {Neural cross-lingual
  named entity recognition with minimal resources}.
\newblock In \emph{Proceedings of the 2018 Conference on Empirical Methods in
  Natural Language Processing}, pages 369--379, Brussels, Belgium. Association
  for Computational Linguistics.

\bibitem[{Xu et~al.(2016)Xu, Reddy, Feng, Huang, and Zhao}]{qa_rel}
Kun Xu, Siva Reddy, Yansong Feng, Songfang Huang, and Dongyan Zhao. 2016.
\newblock \href {https://doi.org/10.18653/v1/P16-1220} {Question answering on
  {F}reebase via relation extraction and textual evidence}.
\newblock In \emph{Proceedings of the 54th Annual Meeting of the Association
  for Computational Linguistics (Volume 1: Long Papers)}, pages 2326--2336,
  Berlin, Germany. Association for Computational Linguistics.

\bibitem[{Xu et~al.(2015)Xu, Mou, Li, Chen, Peng, and Jin}]{lstm2}
Yan Xu, Lili Mou, Ge~Li, Yunchuan Chen, Hao Peng, and Zhi Jin. 2015.
\newblock Classifying relations via long short term memory networks along
  shortest dependency paths.
\newblock In \emph{Proceedings of the 2015 conference on empirical methods in
  natural language processing}, pages 1785--1794.

\bibitem[{Yeh(2000)}]{randomization}
Alexander Yeh. 2000.
\newblock \href {https://www.aclweb.org/anthology/C00-2137} {More accurate
  tests for the statistical significance of result differences}.
\newblock In \emph{{COLING} 2000 Volume 2: The 18th International Conference on
  Computational Linguistics}.

\bibitem[{Zeng et~al.(2014)Zeng, Liu, Lai, Zhou, and Zhao}]{cnnpop}
Daojian Zeng, Kang Liu, Siwei Lai, Guangyou Zhou, and Jun Zhao. 2014.
\newblock \href {https://www.aclweb.org/anthology/C14-1220} {Relation
  classification via convolutional deep neural network}.
\newblock In \emph{Proceedings of {COLING} 2014, the 25th International
  Conference on Computational Linguistics: Technical Papers}, pages 2335--2344,
  Dublin, Ireland. Dublin City University and Association for Computational
  Linguistics.

\bibitem[{Zhang and Wang(2015)}]{kbp37}
Dongxu Zhang and Dong Wang. 2015.
\newblock Relation classification via recurrent neural network.
\newblock \emph{arXiv preprint arXiv:1508.01006}.

\bibitem[{Zou et~al.(2018)Zou, Xu, Hong, and Zhou}]{gan_cross}
Bowei Zou, Zengzhuang Xu, Yu~Hong, and Guodong Zhou. 2018.
\newblock \href {https://www.aclweb.org/anthology/C18-1037} {Adversarial
  feature adaptation for cross-lingual relation classification}.
\newblock In \emph{Proceedings of the 27th International Conference on
  Computational Linguistics}, pages 437--448, Santa Fe, New Mexico, USA.
  Association for Computational Linguistics.

\end{thebibliography}

\appendix

\end{document}